\documentclass[10pt,twocolumn,letterpaper]{article}

\usepackage[pagenumbers]{cvpr} 

\usepackage[dvipsnames]{xcolor}

\definecolor{cvprblue}{rgb}{0.21,0.49,0.74}
\usepackage[pagebackref,breaklinks,colorlinks,citecolor=cvprblue]{hyperref}

\usepackage{booktabs} 
\usepackage{multirow} 
\usepackage{tabularx} 
\usepackage{adjustbox} 

\usepackage[accsupp]{axessibility} 

\def\paperID{10318} 
\def\confName{CVPR}
\def\confYear{2024}

\title{3D-LFM: Lifting Foundation Model}

\author{%
Mosam Dabhi$^{1}$\hspace{2em}%
L{\'a}szl{\'o} A. Jeni$^{1}$\thanks{*Both authors advised equally.}\hspace{2em}%
Simon Lucey$^{2}$\footnotemark[1]\\[1ex] %
$^{1}$Carnegie Mellon University\hspace{2em}%
$^{2}$The University of Adelaide \\[1ex] %
{\tt\small \href{https://3dlfm.github.io}{3dlfm.github.io}}
}

\begin{document}

\setcounter{figure}{-1}
\twocolumn[{%
\renewcommand\twocolumn[1][]{#1}%
\maketitle
\vspace{-1.04cm}
\begin{center}
    \centering
    \captionsetup{type=figure}
    \begin{subfigure}{0.7\linewidth}
        \includegraphics[width=\linewidth]{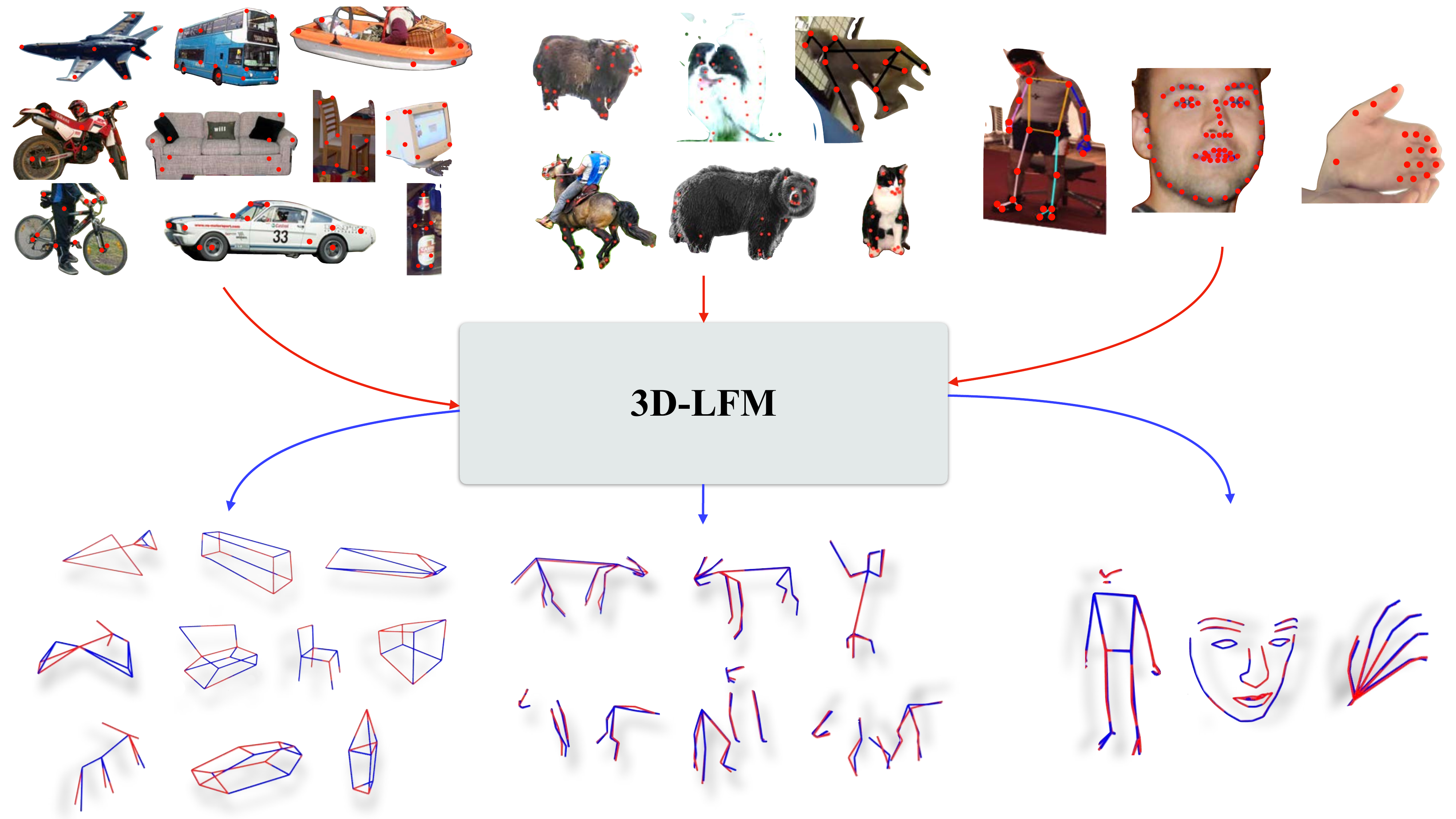}
        \caption{Unified 2D-3D lifting for $30+$ categories.}
        \label{fig:1:overview_sub_1}
    \end{subfigure}%
    \begin{subfigure}{0.3\linewidth}
    \raisebox{0.2\height}{
    \includegraphics[width=\linewidth]{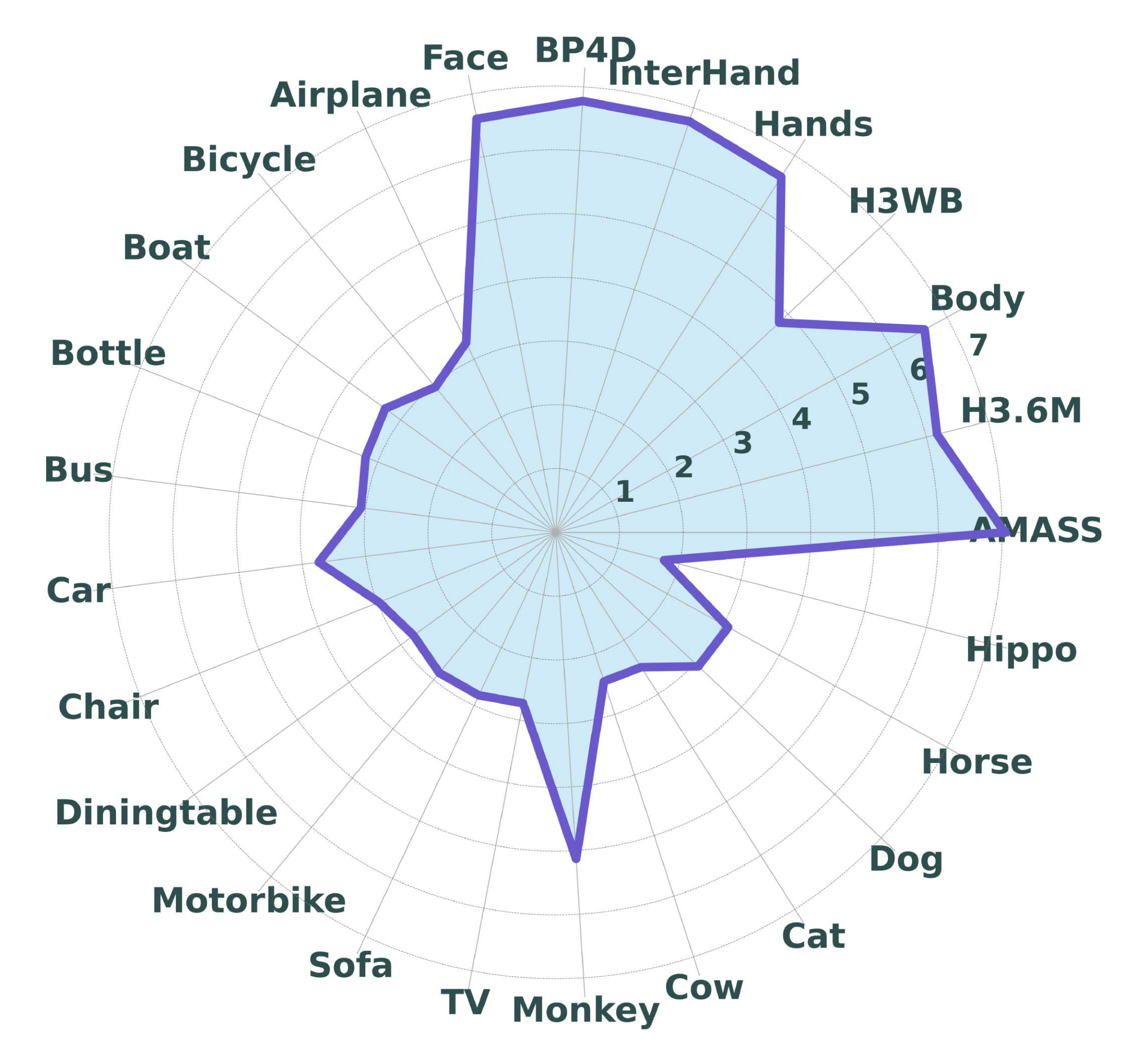} 
    }
        \caption{Dataset diversity visualization.}
        \label{fig:1:overview_sub_radial}
    \end{subfigure}%
    \captionof{figure}{
    \textbf{Overview:} \textbf{(a)} This figure shows the 3D-LFM's ability in lifting 2D landmarks into 3D structures across an array of over 30 diverse categories, from human body parts, to a plethora of animals and everyday common objects. The lower portion shows the actual 3D reconstructions by our model, with red lines representing the ground truth and blue lines showing the 3D-LFM's predictions. \textbf{(b)} This figure displays the model's training data distribution on a logarithmic scale, highlighting that inspite of 3D-LFM being trained on imbalanced datasets, it preserves the performance across individual categories.}
    \label{fig:1}
\end{center}%
}]


\setlength{\parskip}{0pt}


\makeatletter
\renewcommand\@makefnmark{\hbox{\@textsuperscript{*}}}
\makeatother

\footnotetext[1]{Both authors advised equally.}

\begin{abstract}

The lifting of a 3D structure and camera from 2D landmarks is at the cornerstone of the discipline of computer vision. Traditional methods have been confined to specific rigid objects, such as those in Perspective-n-Point (PnP) problems, but deep learning has expanded our capability to reconstruct a wide range of object classes (e.g. C3DPO~\cite{c3dpo} and PAUL~\cite{paul}) with resilience to noise, occlusions, and perspective distortions. However, all these techniques have been limited by the fundamental need to establish correspondences across the 3D training data, significantly limiting their utility to applications where one has an abundance of ``in-correspondence'' 3D data. Our approach harnesses the inherent permutation equivariance of transformers to manage varying numbers of points per 3D data instance, withstands occlusions, and generalizes to unseen categories. We demonstrate state-of-the-art performance across 2D-3D lifting task benchmarks. Since our approach can be trained across such a broad class of structures, we refer to it simply as a 3D Lifting Foundation Model (3D-LFM) -- the first of its kind.

\end{abstract}    
\vspace{-1em}
\section{Introduction}
\label{sec:intro}

Lifting 2D landmarks from a single-view RGB image into 3D has long posed a complex challenge in the field of computer vision because of the ill-posed nature of the problem. This task is important for a range of applications from augmented reality to robotics, and requires an understanding of non-rigid spatial geometry and accurate object descriptions~\cite{deepnrsfm,deepnrsfmpp,bregler2000recovering}. Historically, efforts in single-frame 2D-3D lifting have encountered significant hurdles: reliance on object-specific models, poor scalability, and limited adaptability to diverse and complex object categories. Traditional methods, while advancing in specific domains like human body~\cite{jointformer,motionbert,simplebaseline} or hand modeling~\cite{hand2d3d_1,hand2d3d_2}, often fail when faced with the complexities of varying object types or object rigs (skeleton placements).

To facilitate such single-frame 2D-3D lifting, deep learning methods like C3DPO~\cite{c3dpo} and others~\cite{deepnrsfm,paul,nrsfmformer,mhrnet,deepnrsfmpp} have recently been developed. However, these methods are fundamentally limited in that they must have knowledge of the object category and how the 2D landmarks correspond semantically to the 2D/3D data it was trained upon. Further, this represents a drawback, especially when considering their scaling up to dozens or even hundreds of object categories, with varying numbers of landmarks and configurations. This paper marks a departure from such correspondence constraints, introducing the 3D Lifting Foundation Model (3D-LFM), an object-agnostic single frame 2D-3D lifting approach. At its core, 3D-LFM addresses the limitation of previous models, which is the inability to efficiently handle a wide array of object categories while maintaining high fidelity in 3D keypoint lifting from 2D data. We propose a solution rooted in the concept of permutation equivariance, a property that allows our model to autonomously establish correspondences among diverse sets of input 2D keypoints. 

3D-LFM is capable of performing single frame 2D-3D lifting for $30+$ categories using a single model simultaneously, covering everything from human forms~\cite{h3wb, amass, panopticstudio}, face~\cite{bp4d}, hands~\cite{interhand26m}, and animal species~\cite{acinoset, openmonkeystudio, animal3d}, to a plethora of inanimate objects found in everyday scenarios such as cars, furniture, etc.~\cite{pascal3dplus}. Importantly, 3D-LFM is inherently scalable, poised to expand to hundreds of categories and improve performance, especially in out-of-distribution or less-represented areas, showcasing its broad utility in 3D lifting tasks. 3D-LFM is able to achieve 2D-3D lifting performance that matches those of leading methods specifically optimized for individual categories. The generalizability of 3D LFM is further evident in its ability to handle out-of-distribution (OOD) object categories and rigs, which we refer to as OOD 2D-3D lifting, where the task is to lift the 2D landmarks to 3D for a category never seen during training. We show such OOD results: \textbf{(1)} for inanimate objects - by holding out an object category within the PASCAL dataset, \textbf{(2)} for animals - by training on common object categories such as dogs and cats found in~\cite{animal3d} and reconstructing 3D for unseen and rare species of Cheetahs found in~\cite{acinoset} and in-the-wild zoo captures from~\cite{mbw}, and \textbf{(3)} by showing rig transfer, \textit{i.e.} training 2D to 3D lifting on a Human3.6M dataset rig~\cite{human36m} and showing similar 2D to 3D lifting performance on previously unseen rigs such as those found in Panoptic studio dataaset rig~\cite{panopticstudio} or a COCO dataset rig~\cite{coco}. 3D-LFM transfers learnings from seen data during training to unseen OOD data during inference. It does so by learning general structural features during the training phase via the proposed permutation equivariance properties and specific design choices that we discuss in the following sections.

Recognizing the important role geometry plays in 3D reconstruction~\cite{paul,c3dpo,mbw,mvnrsfm,deepnrsfm,deepnrsfmpp}, we integrate Procrustean methods such as Orthographic-N-Point (OnP) or Perspective-N-Point (PnP) to direct the model’s focus on deformable aspects within a canonical frame. This incorporation significantly reduces the computational burden on the model, freeing it from learning redundant rigid rotations and focusing its capabilities on capturing the true geometric essence of objects. Scalability, a critical aspect of our model, is addressed through the use of tokenized positional encoding (TPE), which, when combined with graph-based transformer architecture, not only enhances the model’s adaptability across diverse categories but also strengthens its ability to handle multiple categories with different number of keypoints and configurations. Finally, the use of skeleton information (joint connectivity) within the graph-based transformers via adjacency matrices provides strong clues about joint proximity and inherent connectivity, aiding in the handling of correspondences across varied object categories.

To the best of our knowledge, 3D-LFM is one of the only known work which is a unified model capable of doing 2D-3D lifting for $30+$ (and potentially even more) categories simultaneously. Its ability to perform unified learning across a vast spectrum of object categories without specific object information and its handling of OOD scenarios highlight its potential as one of the first models capable of serving as a 2D-3D lifting foundation model. 

\noindent The contributions of this paper are threefold:
\begin{figure*} [!h]
    \centering
    \includegraphics[width=0.8\linewidth]{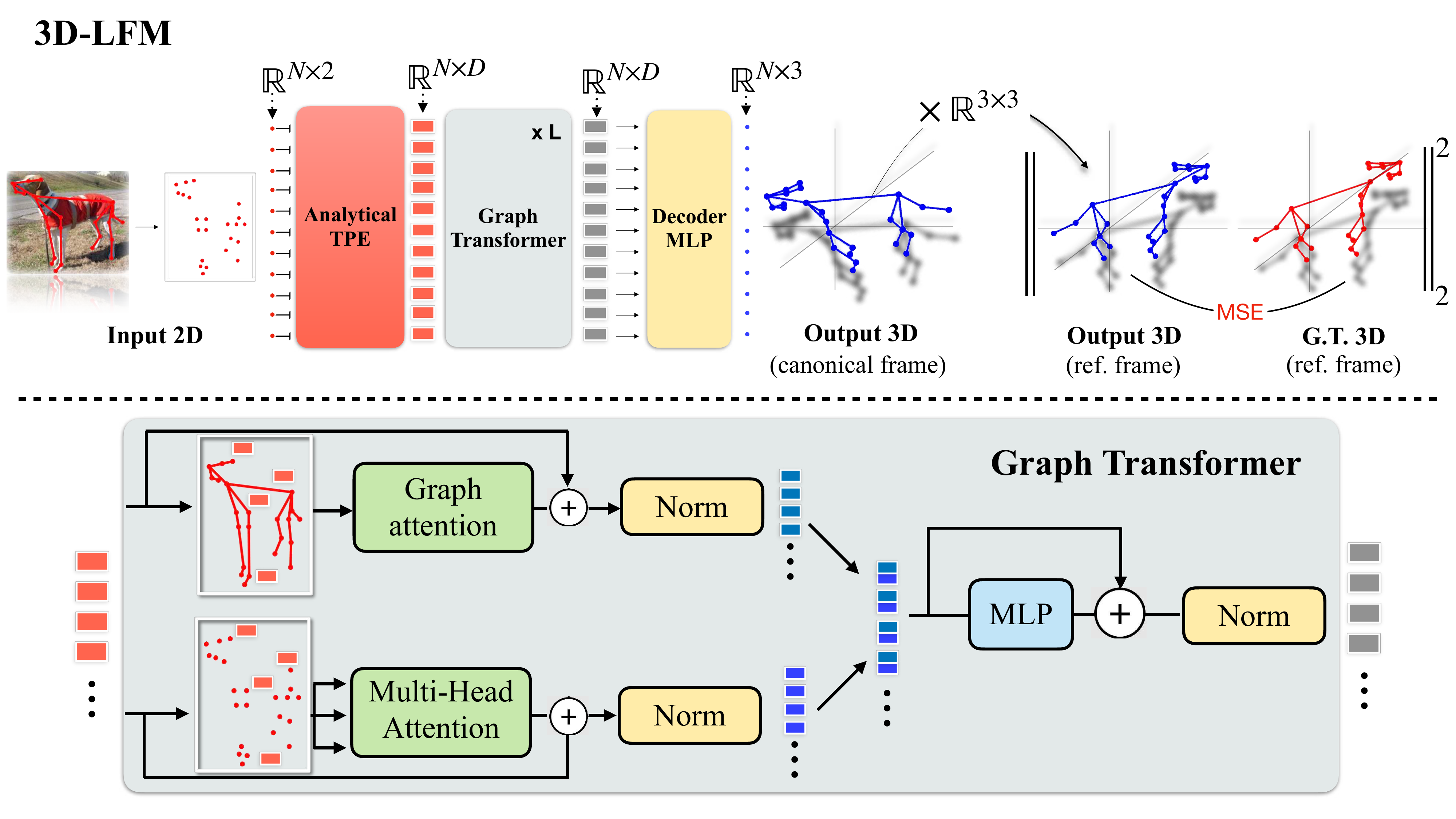}
    \caption{\textbf{Overview of the 3D Lifting Foundation Model (3D-LFM) architecture}: The process begins with the input 2D keypoints undergoing Token Positional Encoding (TPE) before being processed by a series of graph-based transformer layers. The resulting features are then decoded through an MLP into a canonical 3D shape. This shape is aligned to the ground truth (G.T. 3D) in the reference frame using a Procrustean method, with the Mean Squared Error (MSE) loss computed to guide the learning. The architecture captures both local and global contextual information, focusing on deformable structures while minimizing computational complexity.}
    \label{fig:3_method}
\end{figure*}
\begin{enumerate}
    \item We propose a Procrustean transformer that is able to focus solely on learning the deformable aspects of objects within a single canonical frame whilst preserving permutation equivariance across 2D landmarks.
    \item The integration of tokenized positional encoding within the graph-based transformer, to enhance our approach's scalability and its capacity to handle diverse and imbalanced datasets.
    \item We demonstrate that 3D-LFM surpasses state-of-the-art methods in categories such as humans, hands, and faces (benchmark in \cite{h3wb}). Additionally, it shows robust generalization by handling previously unseen objects and configurations, including animals (\cite{acinoset, mbw}), inanimate objects (\cite{pascal3dplus}), and novel object arrangements (rig transfer in \cite{panopticstudio})
\end{enumerate}

In subsequent sections, we explore the design and methodology of our proposed 3D-LFM architecture, including detailed ablation experiments and comparative analyses. Throughout this paper, `keypoints', `landmarks', and `joints' are used interchangeably, referring to specific, identifiable points or locations on an object or figure that are crucial for understanding its structure and geometry.

\section{Related works}
\label{sec:related_works}
The field of 2D-3D lifting has evolved substantially from classic works such as those based on Perspective-n-Point (PnP) algorithms~\cite{epnp}. In these early works, the algorithm was given a set of 2D landmarks and some 3D supervision, namely the known 3D rigid object. The field has since witnessed a paradigm shift with the introduction of deep learning methodologies, led by methods such as C3DPO~\cite{c3dpo}, PAUL~\cite{paul}, and Deep NRSfM~\cite{deepnrsfm}, along with recent transformer-based innovations such as NRSfMFormer~\cite{nrsfmformer}. In these approaches one does not need knowledge of the specific 3D object, instead it can get away with just the 2D landmarks and correspondences to an ensemble of 2D/3D data from the object category to be lifted. However, despite their recent success, all these methods still require that the 2D/3D data be in semantic correspondence. That is, the index to a specific landmark has the same semantic meaning across all instances (e.g. chair leg). In practice, this is quite limiting at run-time, as one needs intimate knowledge of the object category, and rig in order to apply any of these current methods. Further, this dramatically limits the ability of these methods to leverage cross-object and cross-rig datasets, prohibiting the construction of a truly generalizable 2D to 3D lifting foundation model -- a topic of central focus in this paper.


Recent literature in pose estimation, loosely connected to NRSfM but often more specialized towards human and animal body parts, has also seen remarkable progress. Models such as Jointformer~\cite{jointformer} and SimpleBaseline~\cite{simplebaseline} have refined the single-frame 2D-3D lifting process, while generative approaches like MotionCLIP~\cite{motionclip} and Human Motion Diffusion Model~\cite{mdm} have laid the groundwork for 3D generative motion-based foundation models. These approaches, however, are even more limiting than C3PDO, PAUL, etc. in that they are intimately wedded to the object class and are not easily extendable to an arbitrary object class. 




\section{Approach}
\label{sec:method}

Given a set of 2D keypoints representing the projection of an object's joints in an image, we denote the keypoints matrix as $\mathbf{W} \in \mathbb{R}^{N \times 2}$, where $N$ is the predetermined maximum number of joints considered across all object categories. For objects with joints count less than $N$, we introduce a masking mechanism that utilizes a binary mask matrix $\mathbf{M} \in \{0,1\}^{N}$, where each element $m_i$ of $\mathbf{M}$ is defined as:

\begin{equation}
m_i = 
\begin{cases} 
1 & \text{if joint $i$ is present} \\
0 & \text{otherwise}
\end{cases}
\end{equation}

\noindent The 3D lifting function $f: \mathbb{R}^{N \times 2} \rightarrow \mathbb{R}^{N \times 3}$ maps the 2D keypoints to their corresponding 3D structure while compensating for the projection:

\begin{equation}
\mathbf{S} = f(\mathbf{W}) = \mathbf{W} \mathbf{R}^\top + \mathbf{b}
\end{equation}

\noindent where $\mathbf{R} \in \mathbb{R}^{3 \times 3}$ is the projection matrix (assumed either weak-perspective or orthographic) and $\mathbf{b} \in \mathbb{R}^{N \times 3}$ is a bias term that aligns the centroids of 2D and 3D keypoints.

\noindent \textbf{Permutation Equivariance}: To ensure scalability and adaptability across a diverse set of objects, we leverage the property of permutation equivariance inherent in transformer architectures. Permutation equivariance allows the model to process input keypoints $\mathbf{W}$ regardless of their order, a critical feature for handling objects with varying joint configurations:

\vspace{-0.5em}
\begin{equation}
f(\mathcal{P} \mathbf{W}) = \mathcal{P} f(\mathbf{W}) \nonumber
\end{equation}

\noindent where $\mathcal{P}$ is a permutation matrix that reorders the keypoints.

\noindent \textbf{Handling Missing Data:} To address the challenge of missing data, we refer the Deep NRSfM++~\cite{deepnrsfmpp} work and use a masking mechanism to accommodate for occlusions or absences of keypoints. Our binary mask matrix $\mathbf{M} \in \{0,1\}^{N}$ is applied in such a way that it not only pads the input data to a consistent size but also masks out missing or occluded points: $\mathbf{W}_{m} = \mathbf{W} \odot \mathbf{M}$, where $\odot$ denotes element-wise multiplication. To remove the effects of translation and ensure that our TPE features are generalizable, we zero-center the data by subtracting the mean of the visible keypoints:

\vspace{-0.5em}
\begin{equation}
\mathbf{W}_{c} = \mathbf{W}_{m} - \text{mean}(\mathbf{W}_{m})
\end{equation}

We scale the zero-centered data to the range $[-1, 1]$ while preserving the aspect ratio to maintain the geometric integrity of the keypoints. For more details on handling missing data in the presence of perspective effects, we refer the reader to Deep NRSFM++\cite{deepnrsfmpp}.

\noindent \textbf{Token Positional Encoding}: replaces the traditional Correspondence Positional Encoding (CPE) or Joint Embedding which encodes the semantic correspondence information (as used in works such as like~\cite{jointformer,motionbert}) with a mechanism that does not require explicit correspondence or semantic information. Owing to the success of per-point positional embedding, particularly random Fourier features~\cite{robustpct} in handling OOD data, we compute Token Positional Encoding (TPE) using analytical Random Fourier features (RFF) as follows:

\vspace{-0.5em}

\begin{align}
\mathbf{TPE}(\mathbf{W}_{c}) = \sqrt{\frac{2}{D}} \Bigl[ \sin( \mathbf{W}_{c} \boldsymbol{\omega} + b); \cos(\mathbf{W}_{c}\boldsymbol{\omega} + b)\Bigr]
\end{align}

\noindent where $D$ is the dimensionality of the Fourier feature space, $\boldsymbol{\omega} \in \mathbb{R}^{2 \times \frac{D}{2}}$ and $\mathbf{b} \in \mathbb{R}^{\frac{D}{2}}$ are parameters sampled from a normal distribution, scaled appropriately. These parameters are sampled once and kept fixed, as per the RFF methodology. The output of this transformation $\mathbf{TPE}(\mathbf{W}_{c})$ is then fed into the graph-based transformer network as $\mathbf{X}^{\ell}$ where $\ell$ indicates the layer number ($0$ in the above case). This set of features is now ready for processing inside the graph-based transformer layers without the need for correspondence among the input keypoints. The TPE retains the property of permutation equivariance while implicitly encoding the relative positions of the keypoints. 

\subsection{Graph-based Transformer Architecture}

Our graph-based transformer architecture utilizes a hybrid approach to feature aggregation by combining graph-based local attention~\cite{gatpaper}($\mathbf{L}$) with global self-attention mechanisms~\cite{transformerspaper}($\mathbf{G}$) within a single layer (shown as grey block in Fig.~\ref{fig:3_method}. This layer is replicated \( L \) times, providing a sequential refinement of the feature representation across the network's depth.

\noindent \textbf{Hybrid Feature Aggregation:} For each layer \( \ell \), ranging from 0 to \( L \), the feature matrix \( \mathbf{X}^{(\ell)} \in \mathbb{R}^{N \times D} \) is augmented through simultaneous local and global processing. The local processing component, \texttt{GA}($\mathbf{X}^{(\ell)}, \mathbf{A}$), leverages an adjacency matrix \(\mathbf{A}\), which encodes the connectivity based on the object category, to perform graph-based attention on batches of nodes representing the input 2D data:

\vspace{-1em}
\begin{equation}
\begin{split}
\mathbf{L}^{(\ell)} &= \texttt{GA}(\mathbf{X}^{(\ell)}, \mathbf{A}), \\
\mathbf{G}^{(\ell)} &= \texttt{MHSA}(\mathbf{X}^{(\ell)})
\end{split}
\label{eq:hybrid_features}
\end{equation}

\noindent Local and global features are concatenated to form a unified representation \( \mathbf{U}^{(\ell)} \):

\begin{equation}
\mathbf{U}^{(\ell)} = \texttt{concat}(\mathbf{L}^{(\ell)}, \mathbf{G}^{(\ell)})
\end{equation}

\noindent Following the concatenation, each layer applies a normalization(\texttt{LN}) and a multilayer perceptron (\texttt{MLP}). The \texttt{MLP} employs a Gaussian Error Linear Unit \texttt{(GeLU)} as the nonlinearity function to enhance the model's expressive power \begin{equation}
\begin{split}
\mathbf{X}'^{(\ell)} &= \texttt{LN}(\mathbf{U}^{(\ell)}) + \mathbf{U}^{(\ell)}, \\
\mathbf{X}^{(\ell+1)} &= \texttt{LN}(\texttt{MLP}\_{\texttt{GeLU}}(\mathbf{X}'^{(\ell)})) + \mathbf{X}'^{(\ell)}
\end{split}
\label{eq:layer_process}
\end{equation}

Here, \texttt{GA} represents Graph Attention, \texttt{MHSA} denotes Multi-Head Self-Attention, and \texttt{MLP}\_{\texttt{GeLU}} indicates our MLP with GeLU nonlinearity. This architecture is designed to learn patterns in 2D data by considering both the local neighborhood connectivity of input 2D and the global data context of input 2D, which is important for robust 2D to 3D structure lifting.

\subsection{Procrustean Alignment} \label{sec: method: onp}

The final operation in our pipeline decodes the latent feature representation \( \mathbf{X}^{(L)} \) into the predicted canonical structure \( \mathbf{S}_{c} \) via a GeLU-activated MLP:

\[ \mathbf{S}_{c} = \texttt{MLP}_{\text{shape\_decoder}}(\mathbf{X}^{(L)}) \]

\noindent Subsequently, we align \( \mathbf{S}_{c} \) with the ground truth \( \mathbf{S}_{r} \), via a Procrustean alignment method that optimizes for the rotation matrix \( \mathbf{R} \). The alignment is formalized as a minimization problem:

\[ \underset{\mathbf{R}}{\text{minimize}} \quad || \mathbf{M} \odot (\mathbf{S}_{r} - \mathbf{S}_{c}\mathbf{R}) ||_F^2 \]

\noindent where \( \mathbf{M} \) is a binary mask applied element-wise, and \( || \cdot ||_F \) denotes the Frobenius norm. The optimal \( \mathbf{R} \) is obtained via SVD, which ensures the orthonormality constraint of the rotation matrix:


\[ \mathbf{U}, \mathbf{\Sigma}, \mathbf{V}^\top = \text{SVD}((\mathbf{M} \odot \mathbf{S}_{c})^\top \mathbf{S}_{r}), \quad \mathbf{R} = \mathbf{U}\mathbf{V}^\top \]

\noindent The predicted shape is then scaled relative to the reference shape $\mathbf{S}_{r}$, resulting in a scale factor $\gamma$, which yields the final predicted shape $\mathbf{S}_{p}$:

\[ \mathbf{S}_{p} = \gamma \cdot (\mathbf{S}_{c} \mathbf{R}) \]

\noindent This Procrustean alignment step is crucial for directing the model's focus on learning non-rigid shape deformations over rigid body dynamics, thus significantly enhancing the model's ability to capture the true geometric essence of objects by just focusing on core deformable (non-rigid) aspects. The effectiveness of this approach is confirmed by faster convergence and reduced error rates in our experiments, as detailed in Fig.~\ref{fig:tpe-ablation-study}. These findings align with the findings presented in PAUL~\cite{paul}.

\subsection{Loss Function}

The optimization of our model relies on the Mean Squared Error \texttt{(MSE)} loss, which calculates the difference between predicted 3D points \( \mathbf{S}_{p} \) and the ground truth \( \mathbf{S}_{r} \):

\begin{equation}
\mathcal{L}_{\texttt{MSE}} = \frac{1}{N} \sum_{i=1}^{N} \| \mathbf{S}_{p}^{(i)} - \mathbf{S}_{r}^{(i)} \|^2
\end{equation}

Minimizing this loss across \( N \) points ensures the model's ability in reconstructing accurate 3D shapes from input 2D landmarks. This minimization effectively calibrates the shape decoder and the Procrustean alignment to focus on the essential non-rigid characteristics of the objects, helping the accuracy of the 2D to 3D lifting process.

\section{Results and Comparative Analysis}
\label{sec:results}

Our evaluation shows the 3D Lifting Foundation Model (3D-LFM)'s capability in single-frame 2D-3D lifting across diverse object categories without object-specific data in Sec.~\ref{sec: results_multiple_objects_comparison}. Following that, Sec.~\ref{sec: results_object_specific_comparison} highlights 3D-LFM's performance over specialized methods, especially achieving state-of-the-art performance in whole-body benchmarks\cite{h3wb} (Fig. \ref{fig:4.5:sota_comparison}). Additionally, Sec. \ref{sec: results_3D_LFM} shows 3D-LFM's capability in 2D-3D lifting across 30 categories using a single unified model, enhancing category-specific performance and achieving out-of-distribution (OOD) generalization for unseen object configurations during training.
In conclusion, the ablation studies in Section~\ref{sec: results_ablation} validate our proposed procrustean approach, token positional encoding, and the local-global hybrid attention mechanism in the transformer model, confirming their role in 3D-LFM’s effectiveness in both single- and multiple-object scenarios.


\subsection{Multi-Object 3D Reconstruction} \label{sec: results_multiple_objects_comparison}

\textbf{Clarifying naming convention}: In `object-specific' versus `object-agnostic', our primary focus in this naming is on the distinction in \textit{training} methods. Here, object-specific training involves supplying semantic details for each object, leading to isolated training. Conversely, object-agnostic training combines various categories without explicit landmark semantics, leading to combined training.



\noindent \textbf{Experiment Rationale}: 3D-LFM leverages permutation equivariance to accurately lift 2D keypoints into 3D structures across diverse categories, outperforming fixed-array methods by adapting flexibly to variable keypoint configurations. It has been evaluated against non-rigid structure-from-motion approaches~\cite{c3dpo,paul,deepnrsfm,deepnrsfmpp} that require object-specific inputs, showing its ability to handle diverse categories. For a comprehensive benchmark, we utilize the PASCAL3D+ dataset~\cite{pascal3dplus}, following C3DPO's~\cite{c3dpo} methodology, to include a variety of object categories.

\begin{figure}
    \centering
    \includegraphics[width=1\linewidth]{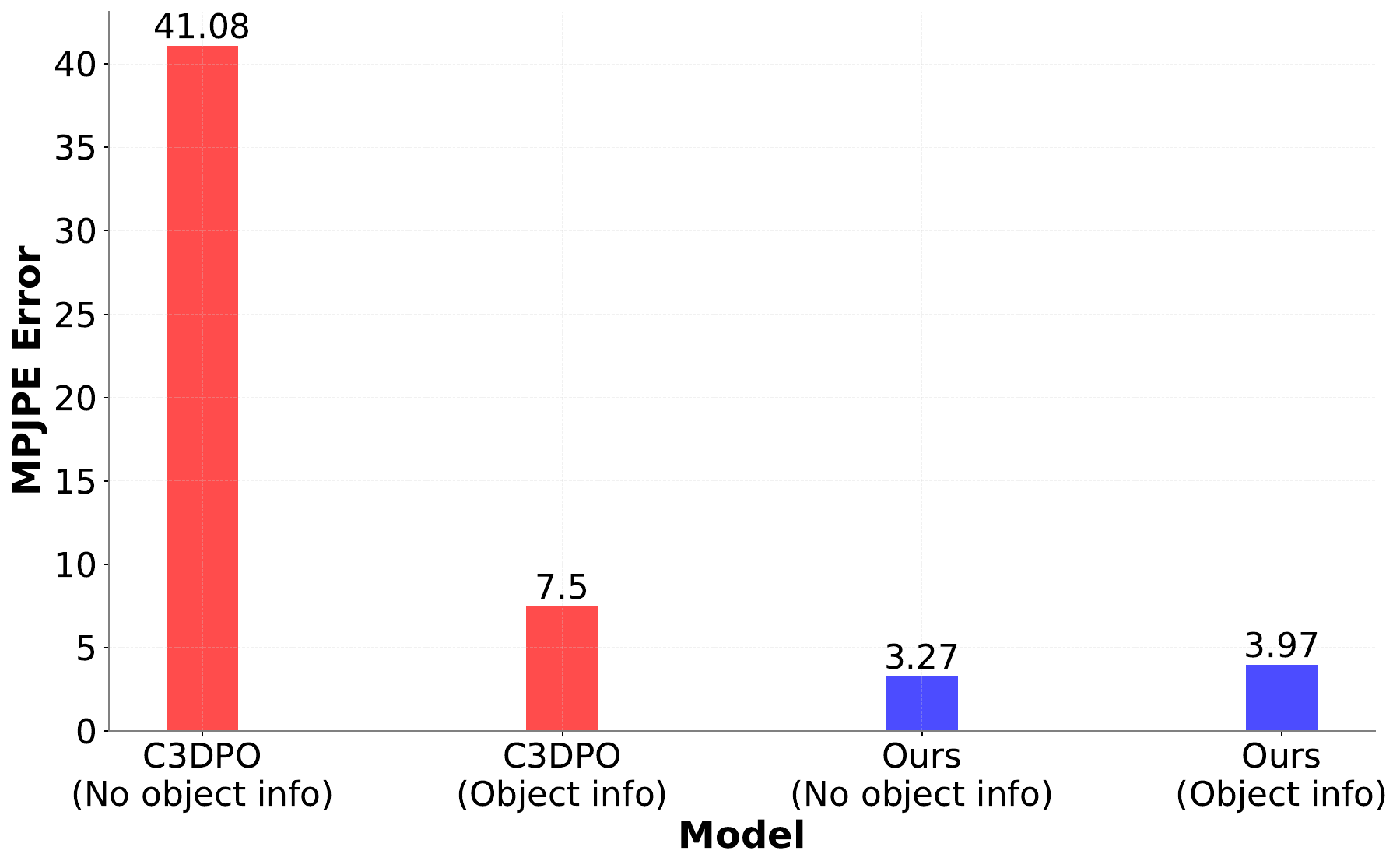}
    \caption{\textbf{3D-LFM vs. C3DPO Performance}: MPJPE comparisons using the PASCAL3D+ dataset, this figure demonstrates our model's adaptability in the absence of object-specific information, contrasting with C3DPO's increased error under the same conditions. The analysis confirms 3D-LFM's superiority across diverse object categories, reinforcing its potential for generalized 2D to 3D lifting.}
    \label{fig:4.1:visual_comparison}    
\end{figure}

\noindent \textbf{Performance:} We benchmark 3D-LFM against the notable NRSfM method, C3DPO~\cite{c3dpo}, for multi-object 2D to 3D lifting with 3D supervision. C3DPO, similar to other contemporary methods~\cite{paul,mhrnet,deepnrsfm,deepnrsfmpp} requiring object-specific details, serves as an apt comparison due to its multi-category approach. Initially replicating conditions with object-specific information, 3D-LFM matches C3DPO's performance, as demonstrated in Fig.~\ref{fig:4.1:visual_comparison}. This stage uses MPJPE to measure 3D lifting accuracy, with C3DPO's training setup including an $MN$ dimensional array for object details where $M$ represents number of objects with $N$ being maximum number of keypoints, and our model is trained separately on each object to avoid providing object-specific information. The 3D-LFM's strength emerges when object-specific data is withheld. While C3DPO shows a decline without such data, 3D-LFM maintains a lower MPJPE across categories, even when trained collectively across categories using only an $N$ dimensional array. These findings (Fig. \ref{fig:4.1:visual_comparison}) highlights 3D-LFM's capabilities, outperforming single-category training and demonstrating its potential as a generalized 2D to 3D lifting solution.


\subsection{Benchmark: Object-Specific Models} \label{sec: results_object_specific_comparison}

\begin{figure}[!t]
    \centering
    \includegraphics[width=1\linewidth]{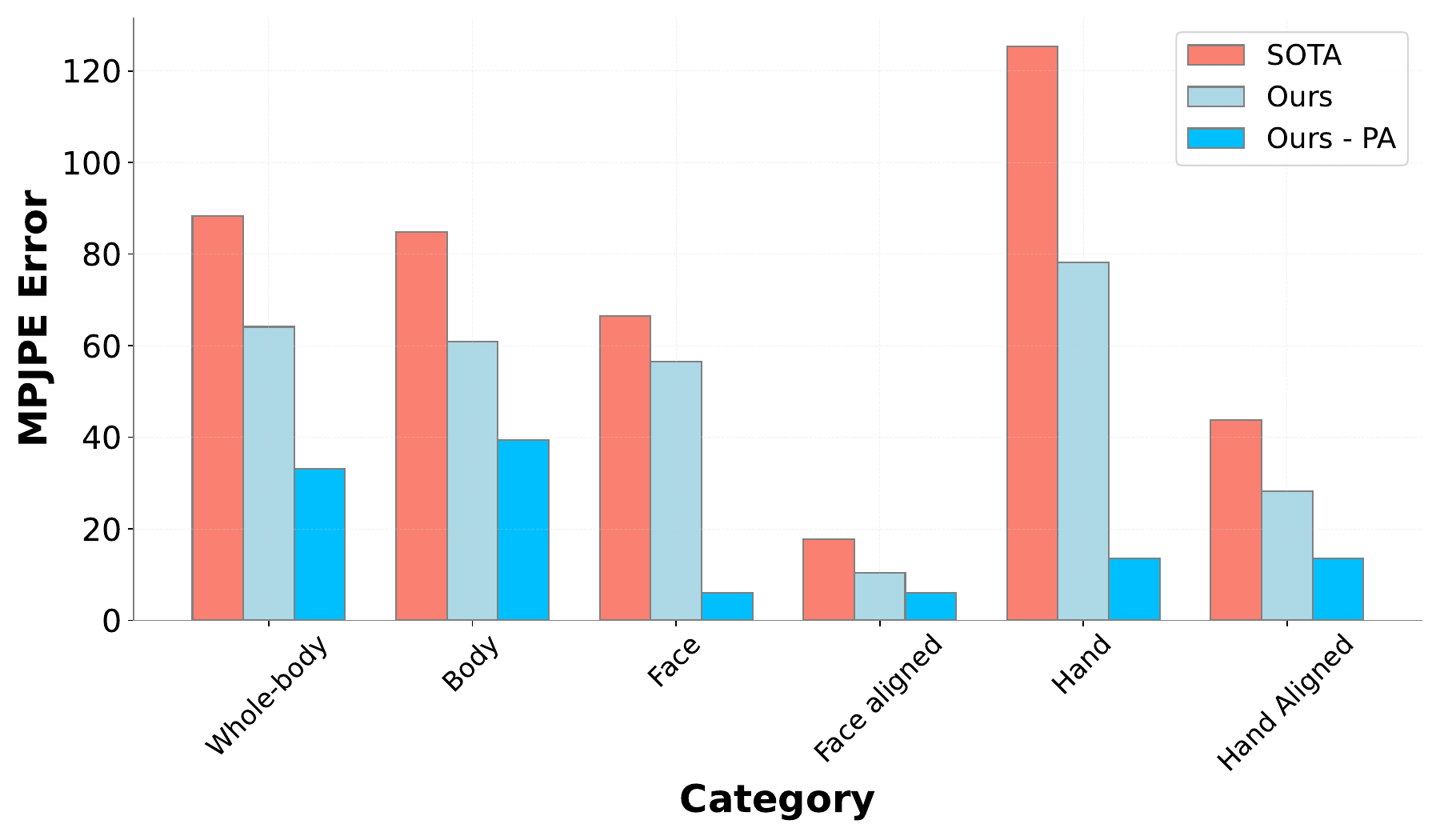}
    \caption{\textbf{Performance Comparison on H3WB Benchmark}: This chart contrasts MPJPE errors for whole-body, body, face, aligned face, hand, and aligned hand categories within the H3WB benchmark~\cite{h3wb}. Our models, with and without Procrustes Alignment (Ours-PA), outperform current state-of-the-art (SOTA) methods, validating our approach's proficiency in 2D to 3D lifting tasks.}
    \vspace{1cm}
    \label{fig:4.5:sota_comparison}
\end{figure}

Next, we benchmark 3D-LFM against leading specialized methods for human body, face, and hands categories. Our model outperforms these specialized methods, showing multi-category learning without the need for category (landmark) semantics. For this study, we evaluate on H3WB dataset~\cite{h3wb}, a recent benchmark for diverse whole-body pose estimation tasks. This dataset is valuable for its inclusion of multiple object categories and for providing a comparative baseline against methods such as Jointformer~\cite{jointformer}, SimpleBaseline~\cite{simplebaseline}, and CanonPose~\cite{canonpose}.  Following H3WB's recommended 5-fold cross-validation and submitting the evaluations to benchmark's authors, we report results on the hidden test set. The results shown in Fig.~\ref{fig:4.5:sota_comparison} and Table~\ref{tab:performance_comparison} include PA-MPJPE and MPJPE, with test set performance numbers provided directly by the H3WB team, ensuring that our results are verified by an independent third-party.

\begin{table}[!t]
\centering
\caption{\textbf{Quantitative performance on H3WB}: Our method demonstrates leading performance across multiple object categories without the need for object-specific designs.}
\label{tab:performance_comparison}
\begin{adjustbox}{width=\columnwidth, center}
\begin{tabular}{@{}lcccc@{}}
\toprule
\textbf{Method} & \textbf{Whole-body} & \textbf{Body} & \textbf{Face/Aligned} & \textbf{Hand/Aligned} \\ \midrule
SimpleBaseline & 125.4 & 125.7 & 115.9 / 24.6 & 140.7 / 42.5 \\
CanonPose w/3D sv. & 117.7 & 117.5 & 112.0 / 17.9 & 126.9 / 38.3 \\
Large SimpleBaseline & 112.3 & 112.6 & 110.6 / 14.6 & 114.8 / 31.7 \\
Jointformer (extra data) & 81.5 & 78 & 60.4 / 16.2 & 117.6 / 38.8 \\
Jointformer & 88.3 & 84.9 & 66.5 / 17.8 & 125.3 / 43.7 \\ \midrule
\textbf{Ours} & \textbf{64.13} & \textbf{60.83} & \textbf{56.55 / 10.44} & \textbf{78.21 / 28.22} \\ 
\textbf{Ours – PA} & \textbf{33.13} & \textbf{39.36} & \textbf{6.02} & \textbf{13.56} \\\bottomrule
\end{tabular}
\end{adjustbox}
\end{table}



\subsection{Towards foundation model} \label{sec: results_3D_LFM}

In this section, we highlight 3D-LFM’s role as a foundational model for varied 2D-3D lifting, capable in managing multiple object types and data imbalances.  In this subsection, we explore 3D-LFM's scalability for collective dataset training (Sec.\ref{sec: combined_dataset}), its generalization to new categories and rig transfer capabilities (Sec.\ref{sec: 4.3.2: ood}). These studies validate the 3D-LFM's role as a foundation model, capable at leveraging diverse data without requiring specific configurations, thus simplifying the 3D lifting process for varied joint setups.

We start this investigation by showing the capability 3D-LFM in handling 2D-3D lifting for $30+$ object categories within the single model, confirming the model's capability to manage imbalanced datasets representative of real-world scenarios as shown in Fig.~\ref{fig:1}. With a comprehensive range of human, hand, face, inanimate objects, and animal datasets, the 3D-LFM is proven to be scalable, without requiring category-specific adjustments. The subsequent subsections will dissect these attributes further, discussing the 3D-LFM's foundational potential in the 3D lifting domain.

\vspace{-1em}
\subsubsection{Combined Dataset Training} \label{sec: combined_dataset}

This study evaluates the 3D-LFM’s performance on isolated datasets against its performance on a combined dataset. Initially, the model was trained separately on animal-based supercategory datasets: specifically \textbf{ OpenMonkey}\cite{openmonkeystudio} and \textbf{Animals3D}\cite{animal3d}.  Subsequently, it was trained on a merged dataset containing a broad spectrum of object categories, including \textbf{Human Body-Based} datasets such as AMASS~\cite{amass} and Human 3.6~\cite{human36m}, \textbf{Hand-Based} datasets such as PanOptic Hands~\cite{panopticstudio}, \textbf{Face-Based} datasets like BP4D+\cite{bp4d}, and various \textbf{Inanimate Objects} from the PASCAL3D+ dataset\cite{pascal3dplus}, along with previously mentioned animal datasets. Isolated training resulted in an average MPJPE of $\mathbf{21.22}$ \textit{mm}, while the combined training method significantly reduced MPJPE to $\mathbf{12.5}$ \textit{mm} on the same animal supercategory validation split. This improvement confirms the potential of 3D-LFM as a pre-training framework and underscores its ability to adapt and generalize from diverse and extensive data collections.


\noindent \textbf{Dataset Selection Rationale:} We selected animal-based supercategory datasets to demonstrate combined training's impact on underrepresented categories. We observed greater performance improvements in smaller, unbalanced datasets (as exemplified by PASCAL3D+: from $\mathbf{4.31}$ \textit{mm} to $\mathbf{1.1}$ \textit{mm} and OpenMonkey: from $\mathbf{19.45}$ \textit{mm} to $\mathbf{9.59}$ \textit{mm}) compared to larger datasets with sufficient balance among categories. Consequently, we see minimal gains in more balanced, larger datasets like AMASS (from $\mathbf{1.67}$ \textit{mm} to $\mathbf{1.66}$ \textit{mm}), underscoring the utility of combined training for enhancing performance in underrepresented and long-tail categories.

\vspace{-1em}

\subsubsection{OOD generalization and rig-transfer:} \label{sec: 4.3.2: ood}

\begin{figure}[!t]
\centering
\includegraphics[width=0.5\textwidth]{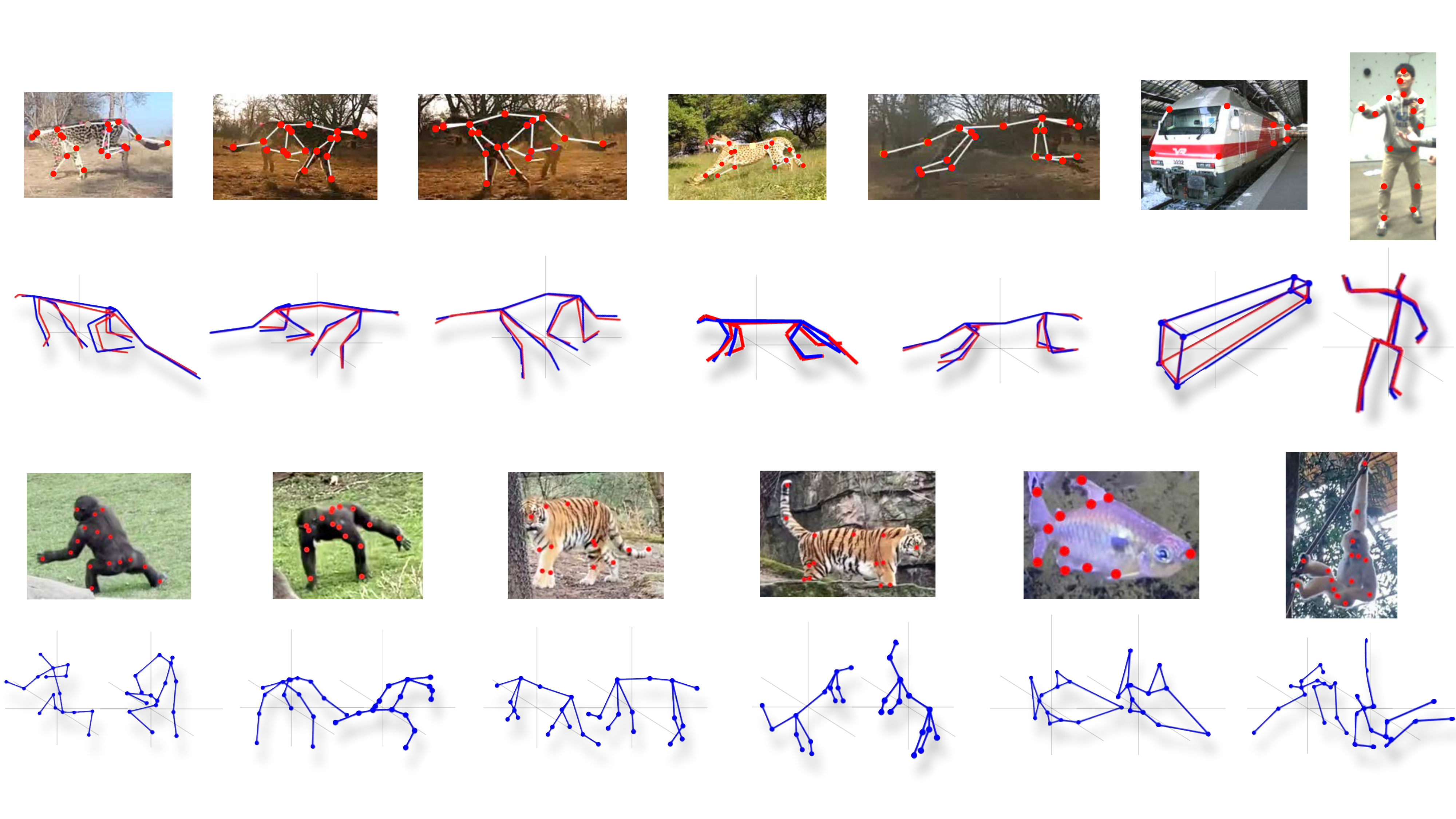}
\caption{\textbf{Generalization to unseen data}: Figure showing 3D-LFM's proficiency in OOD 2D-3D lifting, effectively handling new, unseen categories, and rig generalization from Acinoset~\cite{acinoset} PASCAL3D+~\cite{pascal3dplus}, and Panoptic studio~\cite{panopticstudio} with varying joint arrangements in top row. The bottom row presents in-the-wild data from the MBW dataset~\cite{mbw}, with red dots indicating input keypoints and blue stick figures showing the model’s 3D predictions from different angles.} 
\label{fig:OOD_generalization}
\end{figure}

We evaluate 3D-LFM's generalization to unseen object categories and rig configurations. Its accuracy is highlighted by successful 2D-3D lifting reconstructions of the ``Cheetah'' from Acinoset~\cite{acinoset}, which is not included in the typical Animal3D dataset~\cite{animal3d}, and the ``Train'' category from PASCAL3D+\cite{pascal3dplus}, absent during training. Qualitative reconstructions are shown in Fig. \ref{fig:OOD_generalization}, along with the quantitative results in Tab.\ref{tab:performance} for above categories as well as in-the-wild category like a Chimpanzee from the MBW dataset~\cite{mbw} -- which illustrates model's strong OOD generalization and capability to handle in-the-wild data.

\begin{table}[!t]
\centering
\caption{\textbf{Quantitative evaluation for OOD scenarios}}
\begin{tabular}{@{}lcc@{}}
\toprule
\textbf{Category} & \textbf{OOD} {(mm)} & \textbf{In-Dist.} {(mm)} \\
\midrule
Cheetah~\cite{acinoset} & $26.59$ & $10.16$ \\
Train~\cite{pascal3dplus} & $6.88$ & $5.71$ \\
Chimpanzee~\cite{mbw} & $52.05$ & $42.65$ \\
\bottomrule
\end{tabular}
\label{tab:performance}
\end{table} 



Additionally, we show 3D-LFM's capability in transferring rig configurations between datasets, embodying the concept of generic geometry learning. By training on a $17$-joint Human3.6M dataset~\cite{human36m} and testing on a $15$-joint Panoptic Studio setup~\cite{panopticstudio}, our model gives accurate 3D reconstructions despite variations in joint arrangements. This capability is particularly interesting for its efficiency in utilizing data from multiple rigs of the same object, and underscores the model's adaptability, a cornerstone in processing diverse human datasets. It aligns with the broader community's interest in versatile geometry learning, which makes these findings especially compelling. For a more thorough validation, we direct the reader to the ablation section, where qualitative visuals (Fig. \ref{fig:rig-recon}) and quantitative analysis (Sec. \ref{sec: token_pos_encoding_ablation}) further highlight 3D-LFM's OOD generalization and rig transfer efficacy.
 
\subsection{Ablation} \label{sec: results_ablation}


In our ablation studies, we evaluate the 3D-LFM's design elements and their individual contributions to its performance. Detailed experiments on the Human3.6M benchmark~\cite{human36m} and a blend of other datasets including Animal3D~\cite{animal3d} and facial datasets~\cite{panopticstudio, bp4d} were carried out to ablate the role of Procrustean transformation, hybrid attention mechanisms, and tokenized positional encoding (TPE) in enabling the model's scalability and out-of-distribution (OOD) generalization.

\vspace{-1em}
\subsubsection{Procrustean Transformation} \label{sec: procrustes_ablation}

\begin{figure}[!t]
    \centering
    \begin{subfigure}[b]{0.49\linewidth}
        \centering
        \includegraphics[width=\linewidth]{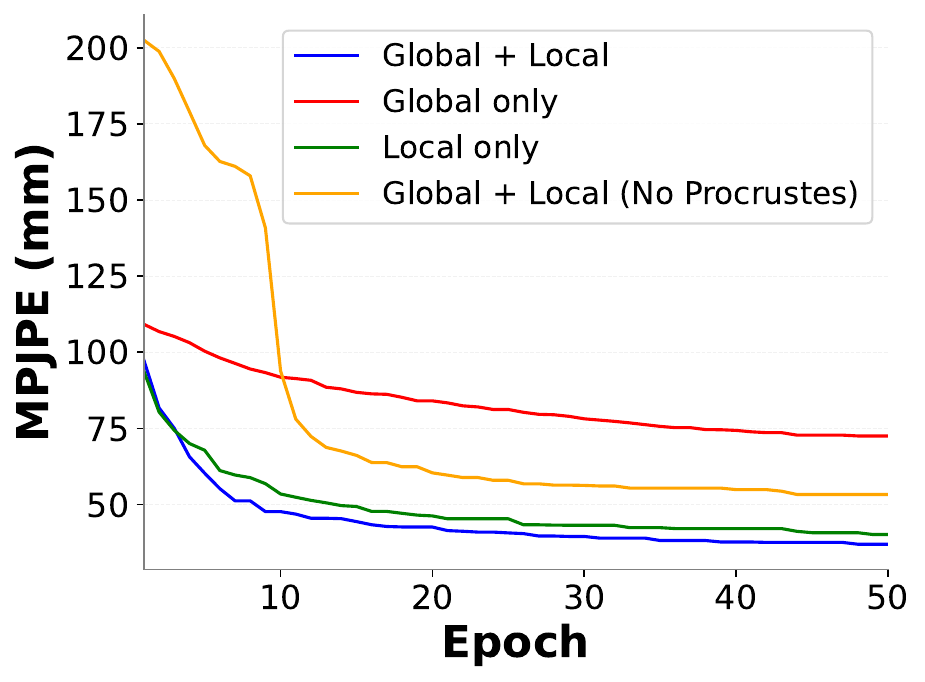}
        \label{fig:all_ablation}
    \end{subfigure}
    \begin{subfigure}[b]{0.49\linewidth}
        \centering
        \includegraphics[width=\linewidth]{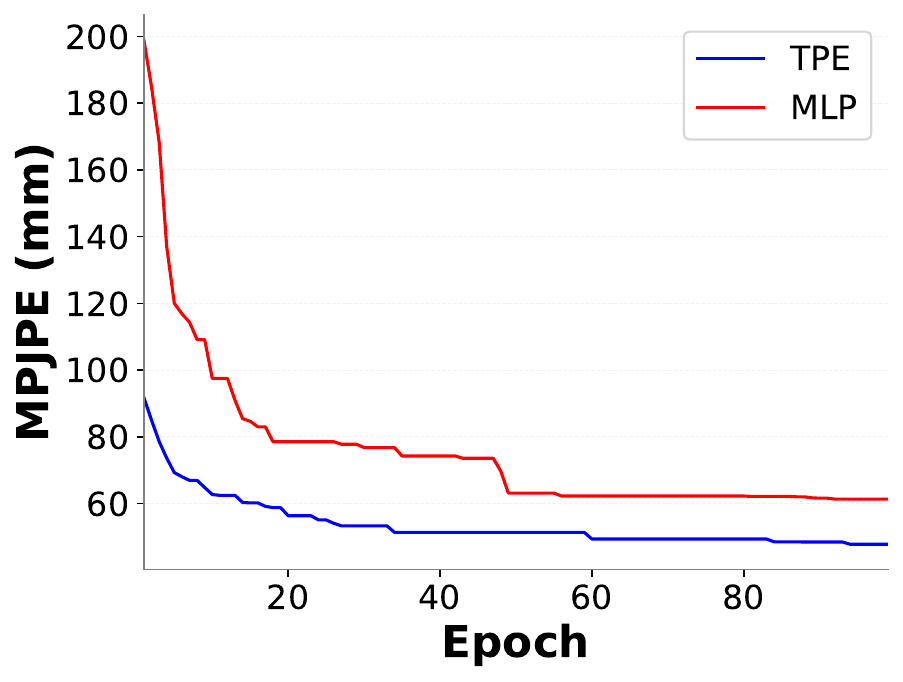}
        \label{fig:tpe_convergence}
    \end{subfigure}     
    \vspace{-1em}
    \caption{\textbf{(a)} Comparing attention strategies in 3D-LFM. The combined local-global approach with procrustean alignment surpasses other configurations in MPJPE reduction over 100 epochs on the Human3.6M validation split. \textbf{(b)} rapid convergence and efficiency of the TPE approach compared to the learnable MLP}
    \label{fig:tpe-ablation-study}
\end{figure}

3D-LFM's fusion of the procrustean approach, a first in transformer-based lifting frameworks, concentrates on deformable object components, as outlined in Sec.\ref{sec: method: onp}. By focusing on shape within a standard canonical reference frame and avoiding rigid body transformations, we see faster learning and a decreased MPJPE, as evident by the gap between blue and orange lines in Fig. \ref{fig:tpe-ablation-study} \textbf{(a)} suggests. This fusion is crucial for learning 3D deformations, while utilizing transformers' equivariance. These findings suggest that even for transformers, avoiding rigid transformations' learning aids convergence, most notably with imbalanced datasets. 


\vspace{-1.3em}

\subsubsection{Local-Global vs. Hybrid Attention} \label{sec: graph_ablation}


In evaluating 3D-LFM's attention strategies, our analysis on the same validation split as above demonstrates the superiority of a hybrid approach combining local (\texttt{GA}) and global (\texttt{MHSA}) attention mechanisms. This integration, particularly when complemented by Procrustean alignment, significantly enhances performance and accelerates convergence, as evidenced in Fig.~\ref{fig:tpe-ablation-study} \textbf{(a)}. The distinct advantage of this hybrid system validates our architectural choices, showcasing its efficiency in reducing MPJPE errors and refining model training dynamics.

\vspace{-1em}
\subsubsection{Tokenized Positional Encoding:} \label{sec: token_pos_encoding_ablation}

\noindent This ablation study assesses the impact of Tokenized Positional Encoding (TPE), which uses analytical Random Fourier Features for encoding positional information. This study examines TPE's influence on model performance in scenarios of data imbalance and rig transfer generalization.

\begin{table}[!t]
\centering
\caption{\textbf{Impact of TPE on Data Imbalance and Rig Transfer}}
\label{tab:TPE_impact}
\begin{adjustbox}{width=\columnwidth, center}
\begin{tabular}{@{}lccc@{}}
\toprule
\textbf{Study} & \textbf{Experiment} & \textbf{Model Size} & \textbf{Improvement (\%)} \\
\midrule
\multirow{3}{*}{Data Imbalance} &\multirow{3}{*}{\shortstack{Underrepr.\\category (Hippo) ~\cite{animal3d}}}  & \multicolumn{1}{c}{$128$} & \multicolumn{1}{c}{$3.27$} \\
                                &                                         & \multicolumn{1}{c}{$512$}  & \multicolumn{1}{c}{$12.28$} \\
                                &                                         & \multicolumn{1}{c}{$1024$} & \multicolumn{1}{c}{$22.02$} \\
\midrule
\multirow{3}{*}{Rig Transfer}   & \multicolumn{1}{c}{$17$~\cite{human36m}- to $15$~\cite{panopticstudio}-joint} &  \multirow{3}{*}{N/A} & \multicolumn{1}{c}{$12$} \\
\multicolumn{1}{c}{}               & \multicolumn{1}{c}{$15$~\cite{panopticstudio}- to $17$~\cite{human36m}-joint} &  & \multicolumn{1}{c}{$23.29$} \\
\multicolumn{1}{c}{}               & \multicolumn{1}{c}{$52$~\cite{panopticstudio}- to $83$~\cite{bp4d}-joint} &  & \multicolumn{1}{c}{$52.3$} \\
\bottomrule
\end{tabular}
\end{adjustbox}
\end{table}

\noindent \textbf{Data imbalance study}: When tested on the underrepresented hippo category from the Animal3D dataset~\cite{animal3d}, TPE based model showed a $3.27$\% improvement in MPJPE over the baseline MLP with a $128$-dimensional model performance as evident in first row of Tab.~\ref{tab:TPE_impact}. This improvement grew with the model size. These results highlight TPE's scalability and its faster convergence, especially relevant in imbalanced, OOD scenarios as detailed in Fig.~\ref{fig:tpe-ablation-study} (b). The observed performance boosts suggest that TPE's analytical nature might be more suited to adapting to novel data distributions. Increasing model size amplifies TPE's benefits, hinting that its fixed analytical approach more adeptly handles OOD intricacies compared to learnable methods like MLPs, which may falter in such situations.

\begin{figure}[!t]
    \centering
    \begin{subfigure}[b]{1\linewidth}
        \centering
        \includegraphics[width=\linewidth]{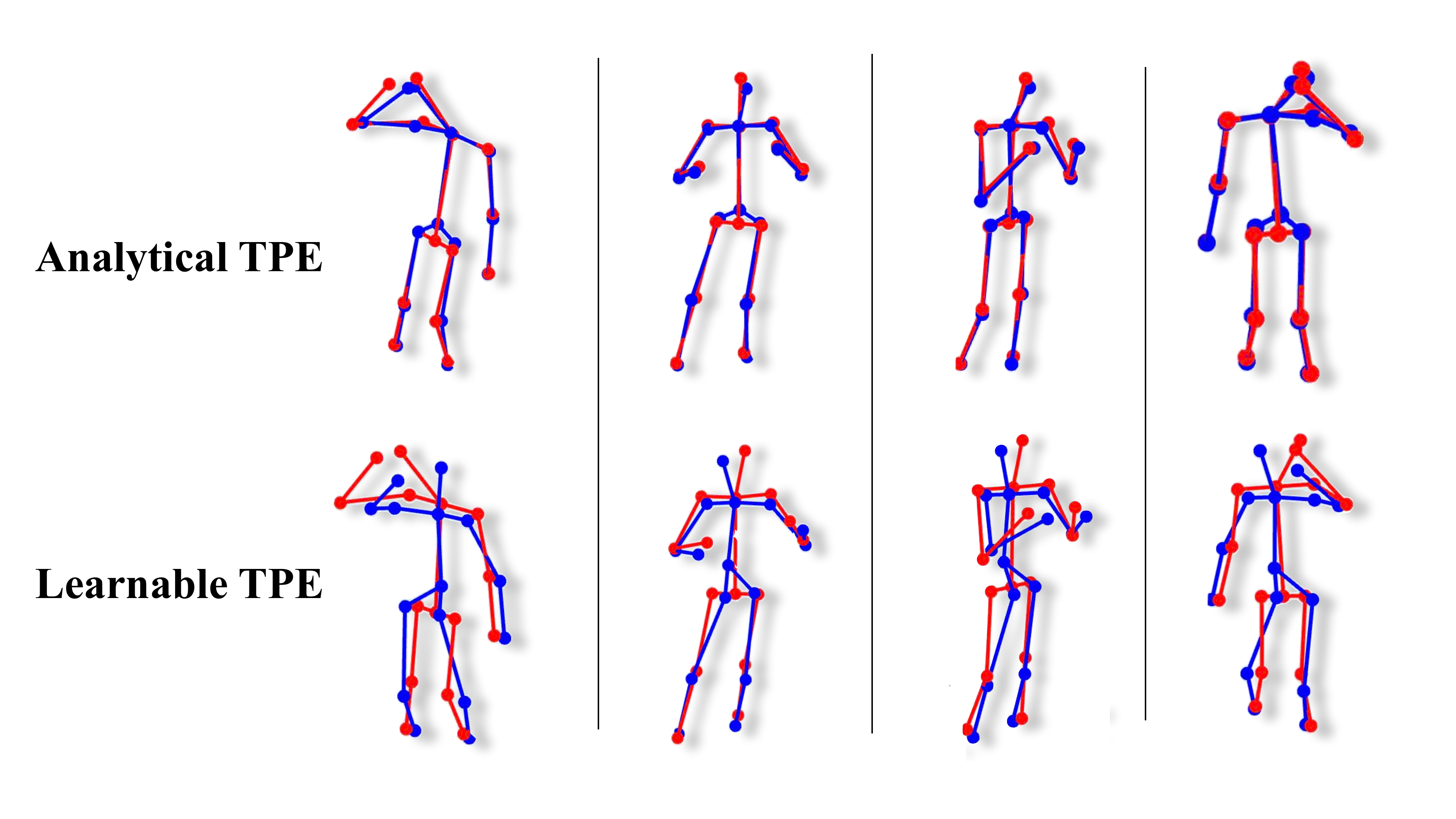}
        \label{fig:rig-recon-subfig}
    \end{subfigure}%
    \vspace{-1em}
    \caption{The qualitative improvement in rig transfer using analytical TPE versus learnable MLP projection. This visualization reinforces the necessity of TPE in handling OOD data such as different rigs, unseen during training.}
    \label{fig:rig-recon}
\end{figure}


\noindent \textbf{Rig transfer study:} Our rig transfer analysis, summarized in Table~\ref{tab:TPE_impact}, showcases TPE's adaptability and effectiveness over the MLP baseline across different joint configurations and rig scenarios, with improvements up to $52.3$\%. These findings, particularly the significant performance boost in complex rig transfers, underscore TPE's robustness in OOD contexts. Figure~\ref{fig:rig-recon} visually highlights the qualitative differences between TPE and MLP approaches in a rig transfer scenario, where the model trained on a $17$-joint~\cite{human36m} configuration is tested on a $15$ joint~\cite{panopticstudio} setup.

\section{Discussion and Conclusion}
\label{sec:discussion}

\vspace{-1em}

The proposed 3D-LFM marks a significant leap in 2D-3D lifting, showcasing scalability and adaptability, addressing data imbalance, and generalizing to new data categories. Its cross-category knowledge transfer requires further investigation and handling of inputs with different perspectives could act as potential limitations. 3D-LFM's efficiency is demonstrated by achieving results comparable to leading methods on~\cite{h3wb} benchmark as well as its proficiency in out-of-distribution (OOD) scenarios on limited computational resources. For training duration and computational details, please refer to the supplementary materials. This work establishes a baseline framework for future 3D pose estimation and 3D reconstruction models. In summary, the 3D-LFM creates a universally applicable model for 3D reconstruction from 2D data, paving the way for diverse applications that requires accurate 3D reconstructions from 2D inputs.

\section*{Acknowledgements}
This research was supported in part by the Australian Research Council under Discovery Project DP220103803.

\maketitlesupplementary
\renewcommand{\thesection}{\Roman{section}}
\setcounter{section}{0}

\section{Training Details}

The 3D Lifting Foundation Model (3D-LFM), as detailed in Sec.~\ref{sec: combined_dataset}, was trained across $30$ diverse categories on a single \texttt{NVIDIA A100 GPU}. This dataset consisted of over $18$ million samples, with data heavily imbalanced and mostly dominated by human datasets as shown in Fig.~\ref{fig:1}. This training highlights the model's practicality, with mixed datasets having imbalance within them. 

\noindent \textbf{Model parameters}: In the architecture of 3D-LFM, the transformer block consists of four layers, with the hidden dimensions and head counts tailored to the dataset scale. Specifically, for datasets exceeding $10,000$ frames, we used a model dimension of $512$ and a head count of $8$. Datasets with frame counts ranging from $1,000$ to $10,000$ were assigned model dimensions of $256$ with $4$ heads. For smaller datasets, consisting $1$ to $1,000$ frames, we employed a more compact model with dimensions set at $64$ and head counts maintained at $4$. This gradation in model complexity ensures a balanced approach, aligning the model capacity with the dataset size, which is particularly critical for achieving computational efficiency and avoiding overfitting on smaller datasets.

\noindent \textbf{Optimizer and scheduler}: \texttt{GeLU} activations were employed for non-linearity in the feedforward layers. The training process was guided by a \texttt{ReduceLROnPlateau} scheduler with a starting learning rate of \texttt{0.001} and a patience of $20$ epochs. An early stopping mechanism was implemented, halting training if no improvement in MPJPE was noted for $30$ epochs, ensuring efficient and optimal performance. This training approach enabled 3D-LFM to surpass leading methods in 3D lifting task proposed by H3WB benchamark~\cite{h3wb}.

\section{Limitations}
\begin{figure}[!t]
    \centering
    \begin{subfigure}[b]{1\linewidth}
        \centering
        \includegraphics[width=\linewidth]{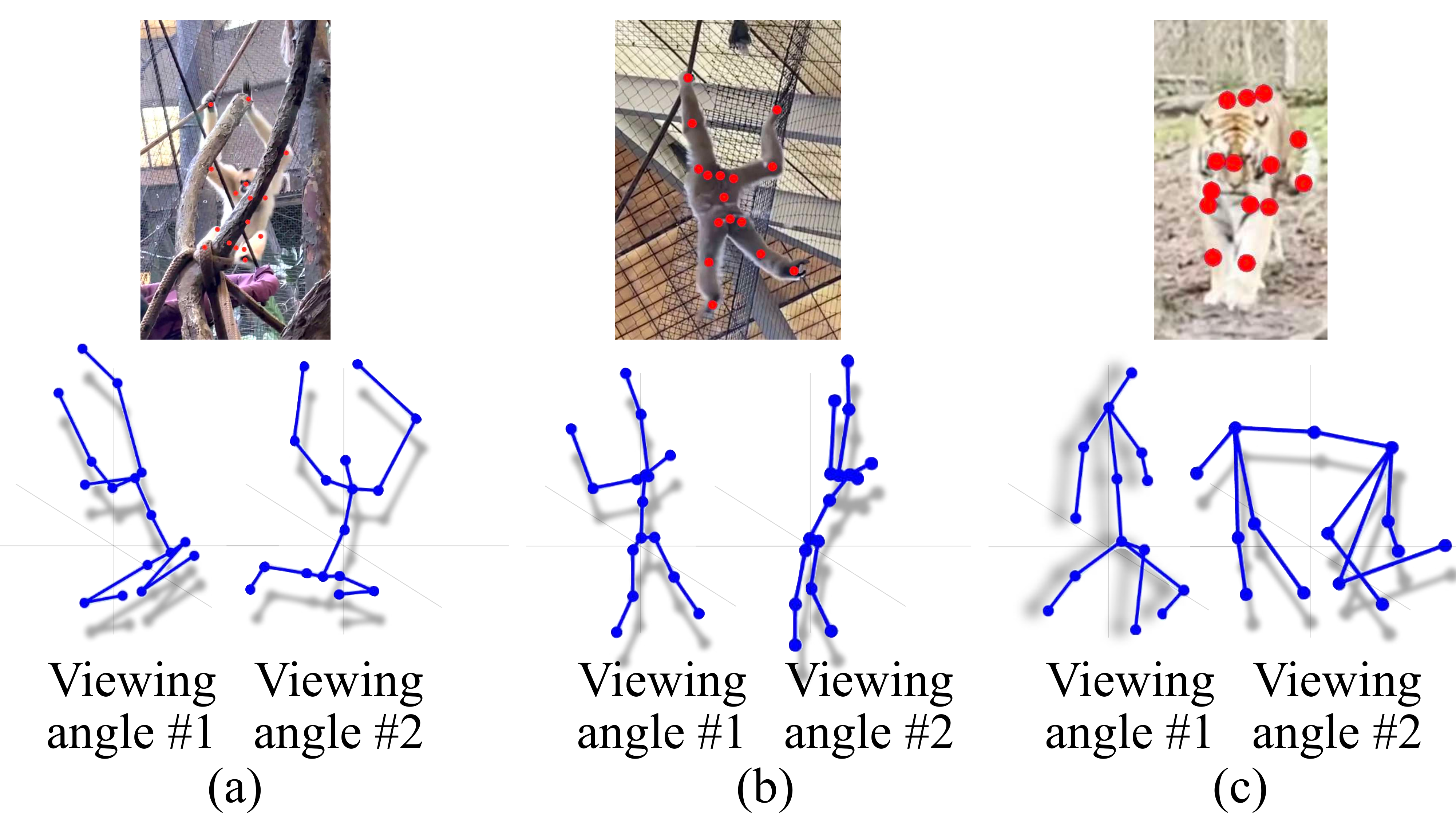}
        \label{fig:rig-recon-subfig}
    \end{subfigure}%
    \vspace{-1em}
    \caption{\textbf{Challenges in Perspective and Depth Perception}: \textbf{(a)} Incorrect leg orientation due to depth ambiguity in monkey capture. \textbf{(b)} Misinterpreted head position in a second monkey example. \textbf{(c)} A tiger's keypoints distorted by perspective, leading to primate-like 3D predictions."}
    \label{fig:limitations}
\end{figure}

\noindent \textbf{Perspective-Induced Misinterpretations}: The 3D-LFM demonstrates a significant capability in generalizing across object categories. However, it can encounter difficulties when extreme perspective distortions cause 2D inputs to mimic the appearance of different categories. For example, unusual viewing angles can cause a tiger to be misconstrued as a primate, illustrated in Fig. \ref{fig:limitations} \textbf{(c)}, due to the similarity in 2D keypoint configurations. Similarly, depth ambiguities can result in incorrect interpretations of spatial arrangements, such as a monkey's leg extending backward instead of forward (Fig. \ref{fig:limitations} \textbf{(a)}) or misperceiving the orientation of a monkey's head (Fig. ~\ref{fig:limitations} \textbf{(b)}). These limitations highlight the complexities of single-frame 2D to 3D lifting, where the model's reliance on geometric keypoint arrangements can be deceptive under certain perspectives. Enhanced depth cues and perspective-aware mechanisms are necessary for more accurate single-view 3D reconstructions, pointing towards future directions to integrate additional appearance or visual features based contextual information for resolving these ambiguities. \label{sec: depth_amb}

\section{Frequently Asked Questions}

\textbf{Q: How does 3D-LFM handle occlusions?}
\newline\textbf{\textit{A:}} 3D-LFM utilizes a masking strategy alongside Tokenized Positional Encoding (TPE) and joint connectivity to effectively manage occlusions. The model's design allows for accurate 3D predictions and reliable category differentiation even when keypoints are obscured. 

\noindent \textbf{Q: Can the 3D-LFM distinguish between different categories under heavy occlusion?}
\newline\textbf{\textit{A:}} Yes, 3D-LFM is designed to distinguish between various categories, such as animals and inanimate objects, even under significant occlusion. This robustness is demonstrated in Fig.~\ref{fig:1} and Fig.~\ref{fig:OOD_generalization}, where the model performs reliably despite the occluded landmarks.

\noindent \textbf{Q: At what point does occlusion begin to affect the model's category identification accuracy?}
\newline\textbf{\textit{A:}} While the model is quite robust to occlusions, there is a threshold beyond which the accuracy begins to diminish. Our ablation study indicates that when more than 60\% of landmarks are occluded, the model's ability to accurately identify object categories is compromised due to insufficient data for the TPE and joint connectivity to operate effectively.

\vspace{3em}
\noindent \textbf{Q: Does 3D-LFM use visual features from images for 3D reconstruction?}
\newline\textbf{\textit{A:}} No, 3D-LFM is focused on reconstructing 3D structures from 2D landmarks and does not directly use visual image features. This design choice streamlines the model's training and application to various object categories without the need for visual contextual information.

\noindent \textbf{Q: How does 3D-LFM handle size variations across different object categories?}
\newline\textbf{\textit{A:}} 3D-LFM predicts structures within a canonical frame, where size variations are managed by Procrustean loss. This allows the model to normalize scale differences and ensures consistent 3D predictions across a wide range of object sizes, from large vehicles to smaller animals.

\noindent \textbf{Q: Is 3D-LFM capable of handling sequential inputs, such as videos?}
\newline\textbf{\textit{A:}} While 3D-LFM is primarily designed for single-frame lifting, its architecture does produce stable and coherent outputs over sequences of 2D landmarks. However, it does not explicitly model temporal dynamics, which is an area we aim to explore to improve the model’s performance on video data.

\noindent \textbf{Q: What are the potential future enhancements for 3D-LFM?}
\newline\textbf{\textit{A:}} Future enhancements include integrating visual features and temporal dynamics to enhance depth perception and object category differentiation. This will likely improve the robustness and accuracy of the model in more complex, real-world scenarios as discussed in Sec.~\ref{sec: depth_amb} of the supplementary material.

\section{More Qualitative Results}
Additional qualitative results and project resources can be found on the project's website (\href{https://3dlfm.github.io/}{https://3dlfm.github.io/}). 
{
    \small
    \bibliographystyle{ieeenat_fullname}
    \bibliography{main}
}


\end{document}